\documentclass[conference]{IEEEtran}
\IEEEoverridecommandlockouts
\usepackage{cite}
\usepackage{amsmath,amssymb,amsfonts}
\usepackage{algorithmic}
\usepackage{graphicx}
\usepackage{textcomp}
\usepackage{xcolor}
\usepackage{tabularx}
\usepackage{subcaption}
\usepackage{float}
\usepackage{url}

\def\BibTeX{{\rm B\kern-.05em{\sc i\kern-.025em b}\kern-.08em
    T\kern-.1667em\lower.7ex\hbox{E}\kern-.125emX}}
    
\begin{document}

\title{Time Series Anomaly Detection via Reinforcement Learning-Based Model Selection\\
}

\author{\IEEEauthorblockN{Jiuqi Elise Zhang, Di Wu, Benoit Boulet}
\IEEEauthorblockA{\textit{Department of Electrical and Computer Engineering} \\
\textit{McGill University}\\
Montreal, Canada \\
elise.zhang@mail.mcgill.ca, di.wu5@mcgill.ca, benoit.boulet@mcgill.ca}
}

\maketitle

\begin{abstract}
Time series anomaly detection has been recognized as of critical importance for the reliable and efficient operation of real-world systems. Many anomaly detection methods have been developed based on various assumptions on anomaly characteristics. However, due to the complex nature of real-world data, different anomalies within a time series usually have diverse profiles supporting different anomaly assumptions. This makes it difficult to find a single anomaly detector that can consistently outperform other models. In this work, to harness the benefits of different base models, we propose a reinforcement learning-based model selection framework. Specifically, we first learn a pool of different anomaly detection models, and then utilize reinforcement learning to dynamically select a candidate model from these base models. Experiments on real-world data have demonstrated that the proposed strategy can indeed outplay all baseline models in terms of overall performance.

\end{abstract}


\section{Introduction}
As an integral part of cyber-physical system technologies, the smart grid technology aims to leverage advanced infrastructures, communication networks and computation techniques to improve the safety and efficiency of power grids. With the recent advancements in machine learning, different machine learning techniques have already found applications in this domain, including sustainable energy management~\cite{zhang2022building, wu2018machine}, electric load forecasting~\cite{DBLP:conf/ijcai/LinW21,lin2021spatial,wu2019multiple,wu2017boosting}, electric vehicle charging forecasting~\cite{huang2020ensemble}, and electric vehicle charging scheduling~\cite{xiong2015impact,wu2017two,wu2014neighborhood}. Anomaly detection, being critical for determining the operating status of cyber-physical systems, is also one potential field of application in this domain and will be the main topic of this paper.

Anomalies, or outliers, are defined as “an observation which appears to be inconsistent with the remainder of that set of data”~\cite{barnett1984outliers}, or data points that ``deviate from patterns or distributions of the majority of the sequence''~\cite{ruff2021unifying}. The occurrence of anomalies usually indicates potential risks in the system, i.e. anomalous meter measurements in power grids could suggest malfunctions or possible cyberattacks~\cite{Gunduz2018Analysis, Skopik2014Dealing, tehrani2019cyber}; anomalies in finance time series might signal ``illegal activities such as fraud, risk, identity theft, networks intrusions, account takeover and money laundering''~\cite{anandakrishnan2018anomaly}. Anomaly detection is, therefore, significant for ensuring the system's operational security, and we have seen applications in fields such as medical systems~\cite{chuah2007ecg,keogh2006finding}, online social networks~\cite{islam2017comprehensive}, Internet of Things (IoT)~\cite{cook2019anomaly}, smart grid~\cite{zhang2021time}, etc.

An anomaly detector is usually built upon one of the following assumptions: 

\textit{(1) Anomalies have a low frequency of occurance.} Under this assumption, if we can characterize the underlying probability distribution of the data, anomalies are most likely to occur in low-probability regions or the tails of a distribution. One work on anomaly detection in data streams~\cite{siffer2017anomaly} fits a distribution of the extreme values of the input data stream based on the Extreme Value Theory (EVT), and utilizes Peaks-Over-Threshold (POT) approach to estimate a normality bound for detecting anomalies. In another work on outlier detection for power system measurements~\cite{lin2019probabilistic}, the authors propose to use kernel density estimation (KDE) to approximate the probabilistic distrubution of the meter measurements, and further propose a probabilistic autoencoder to reconstruct the upper and lower bounds of the statistical intervals of normal data.  

\textit{(2) Anomalous instances are far away from the majority of data, or are most likely to occur in low-density regions.} Methods based on this assumption often incorporate the concept of nearest neighbors and data-specific distance/proximity/density metrics. One recent study~\cite{bergman2020deep} compares detection performance of k-Nearest Neighbors (kNN) algorithm against multiple state-of-the-art deep learning-based self-supervised frameworks, and the authors discover that the simple nearest-neighbor based-approach still outperforms them. In one study on anomaly detection in the Internet of Things (IoT)~\cite{xu2019automatic}, the authors propose a hyperparameter tuning scheme for the Local Outlier Factor (LOF)~\cite{breunig2000lof}, which is a density score calculated based on the distances between neighboring datapoints. 

\textit{(3) If a good representation of the input dataset can be learned, we expect anomalies to have significantly different profiles than normal instances; alternatively, if we further reconstruct or predict future data from the learned representations, we expect the reconstructions based on anomalies to be significantly different from the reconstructions based on normal data.} In a recent work, the authors propose OmniAnomaly~\cite{su2019robust}, an anomaly detection framework comprising GRU-RNN, Planar Normalizing Flows (Planar NF) and a Variational Autoencoder (VAE) for data representation learning and reconstruction. The reconstruction probability under the given latent variables is then used as the anomaly score. Another study on detecting spacecraft anomalies~\cite{hundman2018detecting} uses LSTM-RNN to predict future values of the sequence, and proposes a nonparametric thresholding scheme to interpret the prediction error. Higher prediction error implies higher likeliness of anomalies in the time series.


 Nonetheless, due to the complexity of real-world data, abnormalities can appear in a variety of forms~\cite{chandola2009anomaly}. In the testing phase, an anomaly detector based on a specific assumption about anomalies tends to favors certain aspects over others, i.e. one single model might be sensitive to particular types of anomalies while prone to overlooking others. There is no universal model that can outperform all the other models on different types of input data.

To take advantage of the benefits of multiple base models at different time phases, we propose to dynamically select an optimal detector from the candidate model pool at each time step. In the proposed setting, the current anomaly prediction is based on the output of the currently selected model. Reinforcement learning (RL), as a machine learning paradigm concerned with ``mapping situations into actions by maximizing a numerical reward signal" ~\cite{sutton2018reinforcement}, appears to be a logical solution to the challenge. A reinforcement learning agent aims to learn the optimum decision-making policy by maximizing the total reward. Reinforcement learning has been applied for different types of real-world problems, i.e. electric vehicle charging scheduling~\cite{dang2020ev,dang2019advanced}, home energy management~\cite{zhang2022building,wu2018optimizing}, and traffic signal control~\cite{huang2021modellight}.  It has also demonstrated effectiveness in dynamic model selection in a variety of disciplines, including model selection for short-term load forecasting \cite{feng2019reinforcement}, model combination for time series forecasting \cite{fu2022reinforcement}, and dynamic weighing of an ensemble for time series forecasting \cite{perepu2020reinforcement}.

In this work, we propose a Reinforcement Learning-based Model Selection framework for Anomaly Detection (RLMSAD). We aim to select an optimal detector at each time step based on observations of input time series and predictions of each base model. Experiments on a real-world dataset, \textit{Secure Water Treatment (SWaT)}, show that the proposed framework outperforms each base detector in terms of model precision.

The remainder of the paper is organized as follows. Section~\ref{sec:prob_back} introduces the problem background and the Markov decision process (MDP) settings of the reinforcement learning problem. Section~\ref{sec:meth} describes the proposed model framework. Section~\ref{sec:exp} provides the experimental results on the \textit{Secure Water Treatment (SWaT)} time series. Finally, we conclude the contributions of this work in Section~\ref{sec:con}.

\section{Technical Background}
\label{sec:prob_back}

\subsection{Unsupervised Anomaly Detection in Time Series}
A time series, $\textbf{X}=\{\textbf{x}_1, \textbf{x}_2, ..., \textbf{x}_t\}$, is a sequence of data indexed in time order. It can be either uni-variate, where each $\textbf{x}_i$ is a scalar, or multivariate, where each $\textbf{x}_i$ is a vector. In this paper, the problem of anomaly detection in multivariate time series is addressed in an unsupervised setting. The training sequence $\textbf{X}_{train}$ is a time series with only normal instances, and the testing sequence $\textbf{X}_{test}$ is contaminated with anomalous instances. In the training phase, an anomaly detector is pretrained on $\textbf{X}_{train}$ to capture the characteristics of normal instances. During testing, the detector examines $\textbf{X}_{test}$ and outputs an anomaly score for each instance. Comparing the score with an empirical threshold yields the anomaly label for each test instance.

\subsection{Markov Decision Process and Reinforcement Learning}

Reinforcement learning (RL) is one machine learning paradigm that deals with sequential decision making problems. It aims to train an agent to discover optimal actions in an environment by maximizing the total reward~\cite{sutton2018reinforcement} and is usually modeled as a Markov Decision Process (MDP). The standard MDP is defined as a tuple, $\mathcal{M}= \langle \mathcal{S,A,P,R,\gamma} \rangle$. In this expression, $\mathcal{S}$ is the set of states, $\mathcal{A}$ is the set of actions, $\mathcal{P}(s'|s,a)$ is the matrix of state-transition probability, and $\mathcal{R}(s,a)$ is the reward function. $\gamma$ is the discount factor for reward calculation and is usually between 0 and 1. For deterministic MDPs, each action leads to a certain state, i.e. the state transition dynamic is fixed, so we don't need to consider matrix $\mathcal{P}(s'|s,a)$, and the MDP can in turn be denoted by $\mathcal{M}= \langle \mathcal{S,A,R,\gamma} \rangle$. The return $G_t=\sum_{k=t+1}^T \gamma^{k-t-1} R_k$ is the cumulative future reward after the current time step $t$. A policy $\pi(a|s)$ is a probability distribution maps the current state to the possibility of selecting a specific action. A reinforcement learning agent aims learn a decision policy $\pi(a|s)$ that maximizes the expected total return.

\section{Methodology}
\label{sec:meth}

\subsection{Overall Workflow}

\begin{figure*}[ht!]
    \centering
    \includegraphics[width=0.75\textwidth]{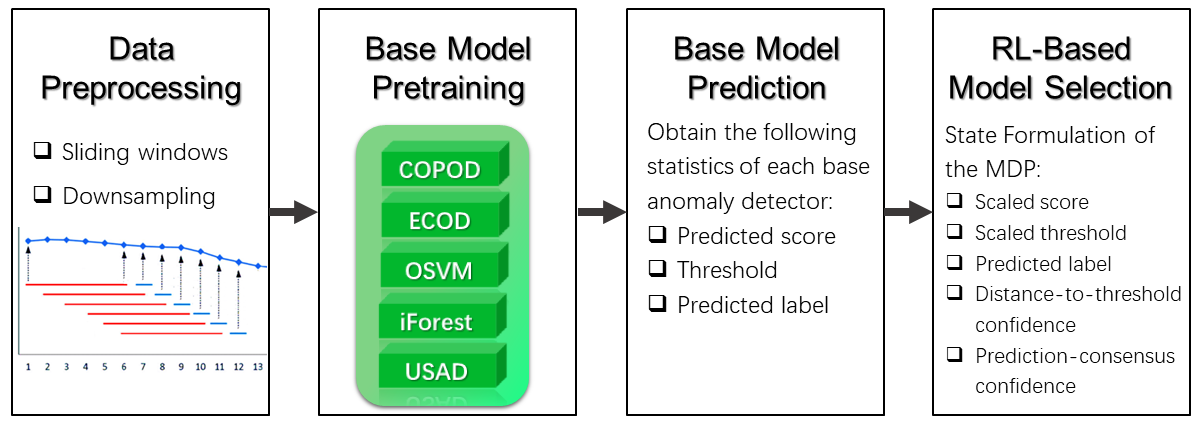}
    \caption{Workflow of our framework: \textit{i)} Data preprocessing; \textit{ii)} Base model pretraining; \textit{iii)} Base model prediction; \textit{iv)} Reinforcement learning for model selection}
    \label{fig:workflow}
\end{figure*}

The workflow of the model is illustrated in Figure \ref{fig:workflow}. The time series input is segmented using sliding windows. In each segmented window, the last timestamp is the target instance for analysis, while all preceding timestamps are used as input features. Each candidate anomaly detector is first pretrained on the training set separately. Then, we run all detectors on the testing data, and each detector computes a sequence of anomaly scores for every testing instance. In order to interpret the anomaly scores, an anomaly threshold needs to be determined empirically for each base detector. Based on the scores and thresholds, all base detectors can then generate their set of predicted labels for the testing data.

Using the predicted scores, empirical thresholds and predicted labels we obtained in the previous step, we define two additional confidence scores \textit{(see Section~\ref{subsec:conf})}. These two confidence scores, along with the predicted scores, thresholds and predicted labels, are then integrated into the Markov decision process (MDP) as state variables \textit{(see Section~\ref{subsec:mdp})}. With the MDP in place, a model selection policy can be learned with appropriate reinforcement learning algorithms (we use DQN in our implementation).

\subsection{Characterizing the Performance of Base Detectors}
\label{subsec:conf}
We propose the following two scores for characterizing the performance of candidate detectors in the model pool. Each anomaly detector produces a sequence of anomaly scores for the testing data. Higher anomaly scores usually indicate higher possibility of anomalies. As a result, the higher a predicted score exceeds the model's threshold, the more likely its corresponding instance is an anomaly under the prediction of the current model. A natural idea would be to characterize the plausibility of a model's prediction by the extent to which a score exceeds the threshold. We hereby propose the \textbf{Distance-to-Threshold Confidence}, \[\frac{Score-Threshold}{Max Score-Min Score}\] which is calculated by the amount by which the current score exceeds the threshold over the difference between the maximum and minimum scores (scaled down to the range of $[0,1]$ by min-max scale).

Inspired by the idea of majority voting in ensemble learning, we also propose the \textbf{Prediction-Consensus Confidence}, \[\frac{\# \ of \ Models \ Predicting \ the \ Same \ Label}{\# \ of \ Models \ in \ the \ Pool}\] which is calculated by the number of models giving the same prediction over the total number of candidate models. The more candidate models that are generating the same prediction within the pool, the more likely the current prediction is true.



    
    
    
    

\subsection{Markov Decision Process (MDP) Formulation}
\label{subsec:mdp}

\begin{figure}[ht!]
    \centering
    \includegraphics[width=0.85\columnwidth]{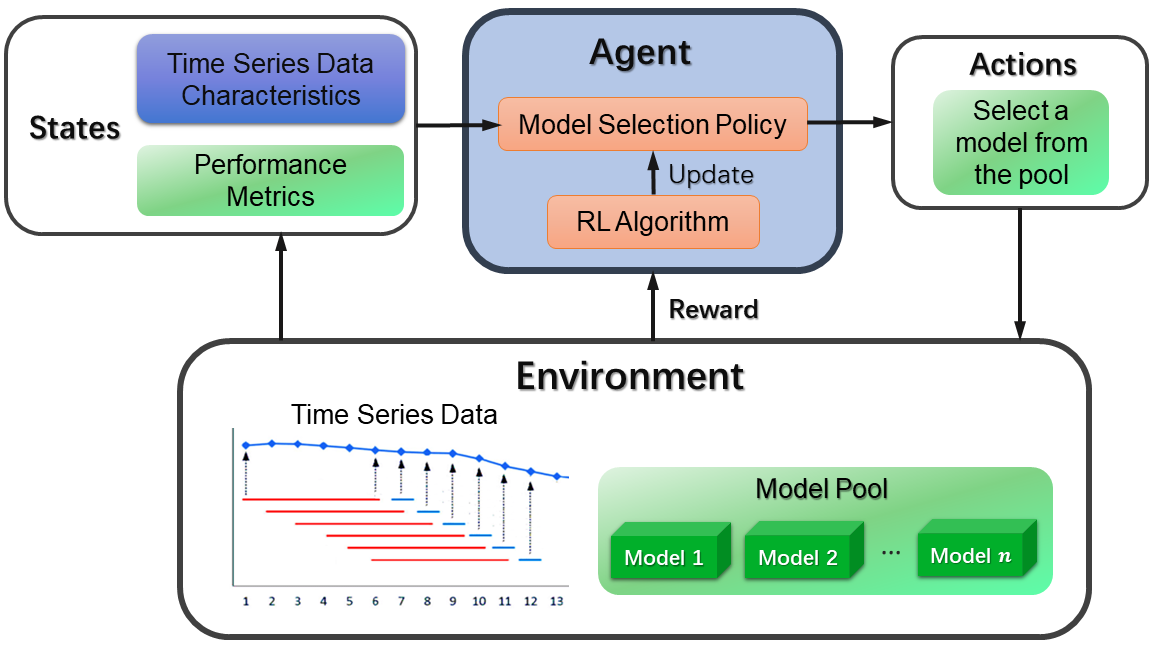}
    \caption{Dynamic selection of anomaly detectors formulated as a reinforcement learning problem}
    \label{fig:rl_env}
\end{figure}

The model selection problem is formalized as a Markov Decision Process in the following manner (see Figure~\ref{fig:rl_env}). In this setting, the state transition naturally follows time order as the time steps in the sequence, and is therefore deterministic, i.e. the state-transition probability $\mathcal{P}(s'|s)$ is 1 for any pair of immediate and consecutive states from $s_t$ to $s_{t+1}$, and 0 for any other cases. 
We set the discount factor $\gamma$ to 1 (no discount).

\begin{itemize}
    \item \textbf{State:}
    The state space is of the same size as the model pool. We consider each state as the selected anomaly detector, and the state variables include the \textit{scaled anomaly score}, the \textit{scaled anomaly threshold}, the binary \textit{predicted label} (0 for normal, 1 for anomalous), the \textit{distance-to-threshold confidence}, and the \textit{prediction-consensus confidence}. Note that for each base detector, its anomaly scores and threshold are normalized to $[0,1]$ using min-max scale.
    \item \textbf{Action:} The action space is discrete and is also of the same size as the model pool. Here, each action of selecting one candidate detector from the pool is characterized by the index of the selected model in the pool.
    \item \textbf{Reward:} The reward function is determined based on the comparison between the predicted label and the actual label. It is denoted by ($r_1, r_2, r_3, r_4 > 0$)
    
        \[R = \left\{ \begin{array}{l}
        {r_1 \ \ \ \ \ \ \ for \ True \ Positive \ (TP)}\\
        {r_2 \ \ \ \ \ \ \ for \ True \ Negative \ (TN)}\\
        {-r_3 \ \ \ \ \ for \ False \ Positive \ (FP)}\\
        {-r_4 \ \ \ \ \ for \ False \ Negative \ (FN)}
        \end{array} \right.\]
    
The reward is designed to encourage the agent to make suitable model selection in a dynamic environment. We regard anomalies as ``positive'', and normal instances as ``negative''. \textit{True positive (TP)} refers to the situation where both the predicted and ground truth labels are ``anomalous'', and the model correctly identifies an anomaly. \textit{True negative (TN)} represents the case where both the predicted and ground truth labels are ``normal'', in other words, the model correctly identifies a normal instance. \textit{False positive (FP)} refers to the case where the predicted label is ``anomalous'' while the ground truth label is ``normal'', suggesting that the model misclassifies a normal instance. \textit{False negative (FN)} represents the scenario where the predicted label is ``normal'' while the ground truth label is ``anomalous'', implying that the model overlooks an anomaly and mistakes it as normal.


    Considering the real-world implications of the above four scenarios, we propose the following assumptions regarding the reward setting: \textit{Normal instances are the majority, and anomalies are relatively rare. Thus, we consider correctly identifying a normal instance as a more trivial case.} In this regard, the reward for TN should be relatively small, while the reward for TP should be relatively large, i.e., $r_1>r_2$.

Also, \textit{neglecting actual anomalies is generally more harmful than giving false alarms.} Sometimes, we would rather have our model be over-sensitive and encourage it to be ``bold'' while making positive predictions. This may cause a model to produce more FPs, but can also reduce the risk of neglecting real anomalies. In this regard, we would penalize the model more severely when it fails to detect an actual anomaly (which is the case of an FN) than when it produces a false alarm (which is the case of an FP), i.e., $r_4>r_3$.

\end{itemize}

\section{Experiments}
\label{sec:exp}
\subsection{Dataset Description}
The dataset used for evaluation is \textit{Secure Water Treatment (SWaT) Dataset} \footnote{\url{https://itrust.sutd.edu.sg/itrust-labs_datasets/dataset_info/}} collected by \textit{iTrust, Centre for Research in Cyber Security} at Singapore University of Technology and Design. This dataset captures the operation data of 51 sensors and actuators within an industrial water treatment testbed. My experiments were conducted on the December 2015 version, which contains 7 days of normal operation data and 4 days of data contaminated by attacks (anomalies). It is a multivariate time series dataset with 51 features. The 7-day normal data contains 496800 timestamps and is selected as the pretraining set. The 4-day data under attack contains 449919 timestamps and is selected as the test set. The percentage of anomalous data in the test set is 11.98\%.

\subsection{Base Models}

When formulating the model pool, we ensure to diversify our selections by choosing models that are based on different anomalies assumptions. The following 5 unsupervised anomaly detection algorithms are selected as candidate models.

\begin{itemize}
    \item \textbf{One-Class SVM} \cite{manevitz2001one} This is a support vector-based method for novelty detection. It aims to learn a hyperplane that separates the high data-density region and the low data-density region.
    \item \textbf{Isolation Forest (iForest)} \cite{liu2008isolation} The Isolation Forest algorithm is based on the assumption that anomalies are more susceptible to isolation. Suppose that multiple decision trees are fitted on the dataset, anomalous data points should be more easily separable from the majority of data. Therefore, we would expect to find anomalies at leaves that are relatively close to the root of a decision tree, i.e., at a more shallow depth of a decision tree.
    \item \textbf{Empirical Cumulative Distribution for Outlier Detection (ECOD)} \cite{li2022ecod} ECOD is a statistical anomaly detection method for multivariate data. It first computes an empirical cumulative distribution along each data dimension, and then utilizes this empirical distribution to estimate the tail probability. The anomaly score is computed by aggregating estimated tail probabilities across all dimensions.
    \item \textbf{Copula-Based Outlier Detection (COPOD)} \cite{li2020copod} COPOD is also a statistical anomaly detection method for multivariate data. It assumes an empirical copula to predict the tail probabilities of all datapoint, which further serves as the anomaly score.
    \item \textbf{Unsupervised Anomaly Detection on Multivariate Time Series (USAD)} \cite{audibert2020usad} This is an anomaly detection method based on representation learning. It learns a robust representation for the raw time-series input using an adversarially trained encoder-decoder pair. During the testing phase, the reconstruction error is used as the anomaly score. The more the score deviates from the expected normal embeddings, the more likely an anomaly has been discovered. 
\end{itemize}

For one-class SVM and iForest, we use \textit{SGDOneClassSVM} and \textit{IsolationForest} with default hyperparameters from the Scikit-Learn library \cite{pedregosa2011scikit}. For ECOD and COPOD, the default functions of \textit{ECOD} and \textit{COPOD} are adopted from the PyOD toolbox \cite{zhao2019pyod}. For USAD, we use the implementation from the authors' original GitHub repository \footnote{\url{https://github.com/manigalati/usad}}.

\subsection{Evaluation Metrics}

In this work, precision (P), recall (R) and F-1 score (F1) are used to evaluate the anomaly detection performance:

\[P=\frac{TP}{TP+FP}, \ R=\frac{TP}{TP+FN}, \ F1=\frac{2\cdot P\cdot R}{P+R}\]

We consider anomalies as ``positive'' and normal data points as ``negative''. By definition, true positives (TP) are correctly predicted anomalies, true negatives (TN) are correctly predicted normal data, false positives (FP) are normal data points wrongly predicted as anomalies, and false negatives (FN) are anomalous data points wrongly predicted as normal.

\subsection{Experimental Settings}
Downsampling can speed up training by reducing the number of timestamps, and can also denoise the overall dataset. We used the same downsampling rate as in paper \cite{audibert2020usad} in the data preprocessing stage. This is conducted by taking an average over every 5 timestamps with a stride of 5. 

Because we already know the percentage of anomalies of the SWaT dataset ($11.98\%$, around $12\%$), the thresholding criterion is fixed at 12\%. This implies that for each base detector, if the score of a data instance ranks among the top 12\% in all the output scores of the sequence, the current detector will label this data instance as an anomaly.

The RL agent is trained using the default settings of \textit{DQN} in a PyTorch-based reinforcement learning toolbox, \textit{Stable-Baselines3}~\cite{raffin2021stable}. For reporting the RL model performance under different hyperparameter settings, we run each experiment with random initialization for 10 times and report the mean and standard deviation of the evaluation metrics.

\subsection{Results and Discussions}

\subsubsection{Overall Performance}
First, we compare the performance of 5 base models alone and the reinforcement learning model selection (RLMSAD) scheme. Here, the reward setting for RLMSAD is TP being $1$, TN being $0.1$, FP being $-0.4$, FN being $-1.5$.


\begin{table}[!htb]
\centering
\begin{tabularx}{0.83\linewidth}{c|ccc}
Model   & \multicolumn{1}{c}{Precision (\%)}                                         & \multicolumn{1}{c}{Recall (\%)}                                         & \multicolumn{1}{c}{F1 (\%)}                                        \\ \hline
iForest  & { 68.14}        & { 64.91}        & { 66.49}        \\
OSVM & 75.39                               & 63.71                               & 69.01                               \\
ECOD    & 66.48                               & 63.33                               & 64.87                               \\
COPOD   & 66.95                               & 63.78                               & 65.33                               \\
USAD    & 71.08                               & 63.86                               & 67.27                               \\ \hline
RLMSAD  & \multicolumn{1}{c}{\textbf{81.05 (4.14)}} & \multicolumn{1}{c}{60.87 (1.24)} & \multicolumn{1}{c}{\textbf{69.45 (1.03)}}
\end{tabularx}
\caption{Performance of base models and the RL Model Selection Scheme}
\label{tab:overall}
\end{table}

The precision scores of the base models range from around 66\% to 75\%. All the base models have demonstrated recall of around 63\%. The F-1 scores of the based models range from around 65\% to 69\%.

Under the proposed framework, the overall precision and F-1 scores have both increased significantly. The precision under RLMSAD has reached 81.05\%, and the F-1 has reached 69.45\%, manifesting substantial improvement in the anomaly detection performance. 

\subsubsection{Different Reward Settings} We investigate the effect of penalization for false positives (FP) and false negatives (FN) in the MDP formulation.

\begin{itemize}
\item \textbf{The Effect of Penalization for False Positives}

 We fix the penalization for false negatives and vary the penalization for false positives. The results are demonstrated in Table \ref{tab:fix_FN}. 

Increasing the penalization for false positives is telling the model to be more prudent about its prediction. Consequently, the model will report fewer false alarms, and will only report an anomaly when it is fairly confident about its prediction. In other words, the model is likely to gain precision by being less sensitive to relatively small deviations within the sequence. It will only report data instances that are significantly different from the majority as anomaly, so that it may cover less actual anomalies and in turn lose recall. This can be demonstrated in Table \ref{tab:fix_FN} by the general reduction in recall score and increase in the precision score as the FP penalization increases.

\begin{table}[!htb]
    \centering
    \scalebox{0.95}{
    \begin{subtable}[ht!]{\columnwidth}
    \centering
        \begin{tabularx}{0.9\columnwidth}{llll}
            \multicolumn{1}{l|}{}                 & Precision (\%)         & Recall (\%)           & F1 (\%)               \\ \hline
            \multicolumn{1}{l|}{FN   $-1$, FP $-0.2$} & 74.69   (3.01) & 62.28   (0.94) & 67.88   (0.96) \\
            \multicolumn{1}{l|}{FN   $-1$, FP $-0.3$} & 78.32   (5.01) & 61.02   (1.06) & 68.49   (1.43) \\
            \multicolumn{1}{l|}{FN   $-1$, FP $-0.4$} & 85.67   (5.80) & 59.70   (1.04) & 70.25   (1.42) \\
            \multicolumn{1}{l|}{FN   $-1$, FP $-0.5$} & 84.94   (6.24) & 59.89   (1.47) & 70.09   (1.30) \\
            \multicolumn{1}{l|}{FN   $-1$, FP $-0.6$}                      & 89.35   (4.90) & 59.33   (1.18) & 71.22   (1.12)
            \end{tabularx}
            \subcaption{Penalization for FN fixed at $-1$, varying penalization for FP}
        \end{subtable}
    }
    \scalebox{0.95}{
    \begin{subtable}[ht!]{\columnwidth}
        \centering
            \begin{tabularx}{.92\columnwidth}{llll}
                \multicolumn{1}{l|}{}                   & Precision (\%)        & Recall (\%)           & F1 (\%)               \\ \hline
                \multicolumn{1}{l|}{FN   $-1.2$, FP $-0.2$} & 79.30   (5.65) & 61.46   (0.95) & 69.12   (1.58) \\
                \multicolumn{1}{l|}{FN   $-1.2$, FP $-0.3$} & 79.70   (5.79) & 61.41   (0.93) & 69.25   (1.63) \\
                \multicolumn{1}{l|}{FN   $-1.2$, FP $-0.4$} & 82.29   (3.95) & 60.80   (0.79) & 69.87   (1.14) \\
                \multicolumn{1}{l|}{FN   $-1.2$, FP $-0.5$} & 83.76   (4.69) & 59.72   (1.10) & 69.64   (0.92) \\
                \multicolumn{1}{l|}{FN   $-1.2$, FP $-0.6$}                      & 85.31   (4.73) & 59.85   (1.19) & 70.26   (0.95)
                \end{tabularx}
                \subcaption{Penalization for FN fixed at $-1.2$, varying penalization for FP}
        \end{subtable}
    }
    \scalebox{0.95}{    
    \begin{subtable}[ht!]{\columnwidth}
        \centering
            \begin{tabularx}{0.92\columnwidth}{llll}
                \multicolumn{1}{l|}{}                   & Precision (\%)        & Recall (\%)           & F1 (\%)               \\ \hline
                \multicolumn{1}{l|}{FN   $-1.5$, FP $-0.2$} & 77.96   (4.62) & 61.54   (0.75) & 68.70   (1.22) \\
                \multicolumn{1}{l|}{FN   $-1.5$, FP $-0.3$} & 75.51   (3.98) & 62.42   (1.08) & 68.27   (1.18) \\
                \multicolumn{1}{l|}{FN   $-1.5$, FP $-0.4$} & 81.05   (4.14) & 60.87   (1.24) & 69.45   (1.03) \\
                \multicolumn{1}{l|}{FN   $-1.5$, FP $-0.5$} & 80.65   (4.54) & 61.25   (1.38) & 69.52   (0.95) \\
                \multicolumn{1}{l|}{FN   $-1.5$, FP $-0.6$}                      & 87.21   (5.01) & 59.94   (1.13) & 70.95   (1.13)
                \end{tabularx}
                \subcaption{Penalization for FN fixed at $-1.5$, varying penalization for FP}
        \end{subtable}
    }   
    \scalebox{0.95}{
    \begin{subtable}[ht!]{\columnwidth}
        \centering
            \begin{tabularx}{0.9\columnwidth}{l|lll}
                 & Precision (\%)        & Recall (\%)           & F1 (\%)               \\ \hline
                FN   $-2$, FP $-0.2$ & 74.36   (4.36) & 62.77   (0.98) & 67.99   (1.39) \\
                FN   $-2$, FP $-0.3$ & 77.12   (2.41) & 61.82   (0.74) & 68.60   (0.89) \\
                FN   $-2$, FP $-0.4$ & 77.86   (4.73) & 61.74   (0.93) & 68.78   (1.23) \\
                FN   $-2$, FP $-0.5$ & 79.94   (5.87) & 61.27(1.40)    & 69.22   (1.47) \\
                FN   $-2$, FP $-0.6$ & 85.36   (5.36) & 60.35   (0.96) & 70.62   (1.45)
                \end{tabularx}
                \subcaption{Penalization for FN fixed at $-2$, varying penalization for FP}
        \end{subtable}
    }
    \caption{Fixing the penalization for FN, varying the penalization for FP}
    \label{tab:fix_FN}
\end{table}

    

\item \textbf{The Effect of Penalization for False Negatives}

We fix the penalization for false positives and vary the penalization for false negatives. The results are demonstrated in Table \ref{tab:fix_FP}.

Increasing the penalization for false negatives is encouraging the model to be ``bolder'' in terms of reporting anomalies. When the cost of producing false negatives (i.e., missing actual anomalies) becomes high, the best strategy for the agent is to report as many anomalies as possible to avoid missing a lot of actual anomalies. In this case, the model will likely lose precision by being over-sensitive to small deviations. On the other hand, since is more likely to flag an instance as anomaly, it is also more likely to cover more actual anomalies and achieve a higher recall. This can be demonstrated by a general decrease in the precision score and an increase in recall score in Table~\ref{tab:fix_FP}.

\begin{table}[!htb]
\centering
\scalebox{0.95}{
\begin{subtable}[ht!]{\columnwidth}
\centering
\begin{tabularx}{.92\columnwidth}{l|lll}
                   & {Precision (\%)} & {Recall (\%)} & {F1 (\%)} \\ \hline
FN   $-1$, FP $-0.2$   & 74.69   (3.01)                & 62.28   (0.94)             & 67.88   (0.96)         \\
FN   $-1.2$, FP $-0.2$ & 79.30   (5.65)                & 61.46   (0.95)             & 69.12   (1.58)         \\
FN   $-1.5$, FP $-0.2$ & 77.96   (4.62)                & 61.54   (0.75)             & 68.70   (1.22)         \\
FN   $-2$, FP $-0.2$   & 74.36   (4.36)                & 62.77   (0.98)             & 67.99   (1.39)        
\end{tabularx}
\subcaption{Penalization for FP fixed at $-0.2$, varying penalization for FN}
\end{subtable}
}
\scalebox{0.95}{
\begin{subtable}[ht!]{\columnwidth}
\centering
\begin{tabularx}{.92\columnwidth}{l|lll}
                   & Precision (\%)         & Recall (\%)            & F1 (\%)                \\ \hline
FN   $-1$, FP $-0.4$   & 85.67   (5.80) & 59.70   (1.04) & 70.25   (1.42) \\
FN   $-1.2$, FP $-0.4$ & 82.29   (3.95) & 60.80   (0.79) & 69.87   (1.14) \\
FN   $-1.5$, FP $-0.4$ & 81.05   (4.14) & 60.87   (1.24) & 69.45   (1.03) \\
FN   $-2$, FP $-0.4$   & 77.86   (4.73) & 61.74   (0.93) & 68.78   (1.23)
\end{tabularx}
\subcaption{Penalization for FP fixed at $-0.4$, varying penalization for FN}
\end{subtable}
}
\scalebox{0.95}{
\begin{subtable}[ht!]{\columnwidth}
\centering
\begin{tabularx}{0.92\columnwidth}{l|lll}
                   & Precision (\%)         & Recall (\%)            & F1 (\%)                \\ \hline
FN   $-1$, FP $-0.6$   & 89.35   (4.90) & 59.33   (1.18) & 71.22   (1.12) \\
FN   $-1.2$, FP $-0.6$ & 85.31   (4.73) & 59.85   (1.19) & 70.26   (0.95) \\
FN   $-1.5$, FP $-0.6$ & 87.21   (5.01) & 59.94   (1.13) & 70.95   (1.13) \\
FN   $-2$, FP $-0.6$   & 85.36   (5.36) & 60.35   (0.96)  & 70.62   (1.45)
\end{tabularx}
\subcaption{Penalization for FP fixed at $-0.6$, varying penalization for FN}
\end{subtable}
}
\caption{Fixing the penalization for FP, varying the penalization for FN}
\label{tab:fix_FP}
\end{table}

    

\end{itemize}

\subsubsection{Ablation Study}

\begin{figure}[!ht]
    \centering
    \includegraphics[width=\columnwidth]{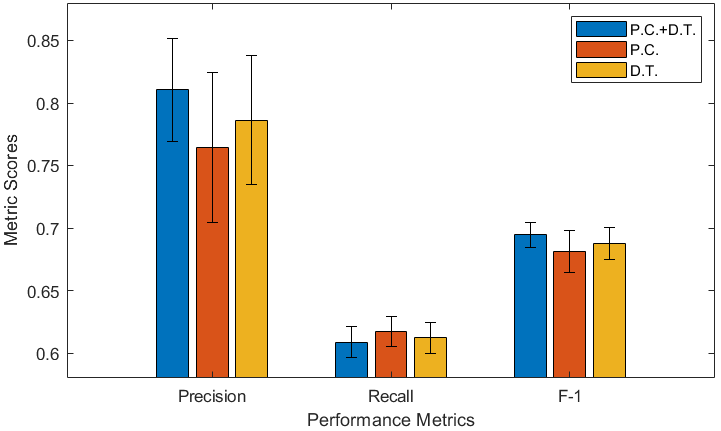}
    \caption{Performance under full environment (P.C.+D.T), environment with only prediction-consensus (P.C.) confidence, and the environment with only distance-to-threshold (D.T.) confidence}
    \label{fig:ablation}
\end{figure}

An ablation study is also conducted to examine the effect of two confidence scores, the \textit{distance-to-threshold (D.T.) confidence} and \textit{prediction-consensus (P.C.) confidence}. Here, two additional reinforcement learning environments are constructed apart from the original one. In these two environments, each one of the two confidence scores is removed from the state variables. We retrain a policy on each of these three environments and report the detection performance in Figure~\ref{fig:ablation}. From Figure~\ref{fig:ablation}, we observe that removal of either of the two confidence scores results in a significant decline in precision and F-1 scores. The optimal performance is achieved only when the two confidence scores are both taken into account as state components.

\section{Conclusions}
\label{sec:con}
In this paper, we proposed a reinforcement learning-based model selection framework for time series anomaly detection. Specifically, two scores, \textit{distance-to-threshold confidence} and \textit{prediction-consensus confidence}, are first introduced to characterize the detection performance of base models. The model selection problem is then formulated as a Markov decision process, with the two scores serving as RL state variables. We aim to learn a model selection policy for anomaly detection in order to optimize the long-term expected performance. Experiments on a real-world time series have been implemented, showcasing the effectiveness of our model selection framework. In the future, we plan to focus on studying adaptive threshold strategy and extracting appropriate data-specific features and embedding them into the state representations of RL frameworks, which may provide the RL agent with more informative state descriptions and potentially result in a more robust performance.

\bibliographystyle{unsrt}
\bibliography{ref}

\end{document}